\documentclass[lettersize,journal]{IEEEtran}
\usepackage{caption}
\usepackage{amsmath,amsfonts}
\usepackage{subcaption}
\usepackage{algorithm}
\usepackage{array}
\usepackage[pagebackref=true,breaklinks,colorlinks,citecolor=blue,linkcolor=blue,urlcolor=blue,hypertexnames=false]{hyperref}
\usepackage{textcomp}
\usepackage{stfloats}
\usepackage{url}
\usepackage{verbatim}

\usepackage{algpseudocode}
\usepackage{booktabs}       % professional-quality tables
\usepackage{nicefrac}       % compact symbols for 1/2, etc.
\usepackage{microtype}      % microtypography
\usepackage{xcolor}         % colors

\usepackage{overpic}

\usepackage{soul}
\usepackage{multirow}

\usepackage{amssymb}
\usepackage{mathtools}
\usepackage{amsthm}
\usepackage{bm}

\newcommand{\answerTODO}[1][]{\textcolor{red}{\bf [TODO]}}

\DeclareRobustCommand\onedot{\futurelet\@let@token\@onedot}

\def\ie{\emph{i.e., }}

\def\etal{\emph{et al.} }
\makeatother

\usepackage{newfloat}
\usepackage{listings}
\DeclareCaptionStyle{ruled}{labelfont=normalfont,labelsep=colon,strut=off} % DO NOT CHANGE THIS
\floatstyle{ruled}
\newfloat{listing}{tb}{lst}{}
\floatname{listing}{Listing}

\begin{document}

\title{PromptRR: Diffusion Models as Prompt Generators for Single Image Reflection Removal}

\author{Tao Wang,
        Wanglong Lu,
        Kaihao Zhang,
        Tong Lu,
        Ming-Hsuan Yang%~\IEEEmembership{Fellow,~IEEE}
        \\
        
% <-this % stops a space
\IEEEcompsocitemizethanks{\IEEEcompsocthanksitem T. Wang and T. Lu are with the State Key Laboratory for Novel Software Technology, Nanjing University, Nanjing, China (e-mail: taowangzj@gmail.com, lutong@nju.edu.cn).
\IEEEcompsocthanksitem W. Lu is with the Department of Computer Science at Memorial University of Newfoundland, St. John's, Canada (e-mail: wanglongl@mun.ca).  
\IEEEcompsocthanksitem K. Zhang is with Harbin Institute of Technology (Shenzhen), Shenzhen, China (e-mail: super.khzhang@gmail.com).
\IEEEcompsocthanksitem M.-H. Yang is with the School of Engineering, University of California at Merced, Merced, CA, USA 
(e-mail: mhyang@ucmerced.edu).}% <-this % stops an unwanted space
%\thanks{Manuscript received XXX; revised XXX.}
}


\maketitle

\begin{abstract}
Existing deep learning-based single-image reflection removal (SIRR) methods often fail to effectively capture key low-frequency (LF) and high-frequency (HF) differences in images, limiting their ability to remove reflections.
To address this limitation, we propose a novel prompt-guided reflection removal framework, PromptRR, which leverages frequency information as visual prompts to enhance reflection removal performance. Specifically, our framework decomposes the reflection removal process into two stages: prompt generation and prompt-guided restoration. In the prompt generation stage, we introduce a prompt pre-training strategy to train a frequency prompt encoder that encodes ground-truth images into LF and HF prompts. Subsequently, we employ diffusion models (DMs) as prompt generators to estimate these prompts based on the pre-trained frequency prompt encoder.
For the prompt-guided restoration stage, we integrate the generated frequency prompts into PromptFormer, a novel Transformer-based network. We further design a specialized prompt block to guide the model towards improved reflection removal effectively.
Extensive experiments on widely used benchmark datasets demonstrate that our approach outperforms state-of-the-art methods. The codes and models are available at \url{https://github.com/TaoWangzj/PromptRR}.

\end{abstract}

\section{Introduction}
\label{sec:intro}

Images taken through transparent surfaces, such as glass, usually contain undesirable specular reflections, significantly compromising the quality and visibility of the captured scenes. This degradation notably impacts the performance of computer vision applications like object detection and face recognition. Thus, single image reflection removal (SIRR) has received considerable interest in recent years.

\begin{figure}[t]
\begin{center}
 % \vspace{-4mm}
	\includegraphics[width=0.5\textwidth]{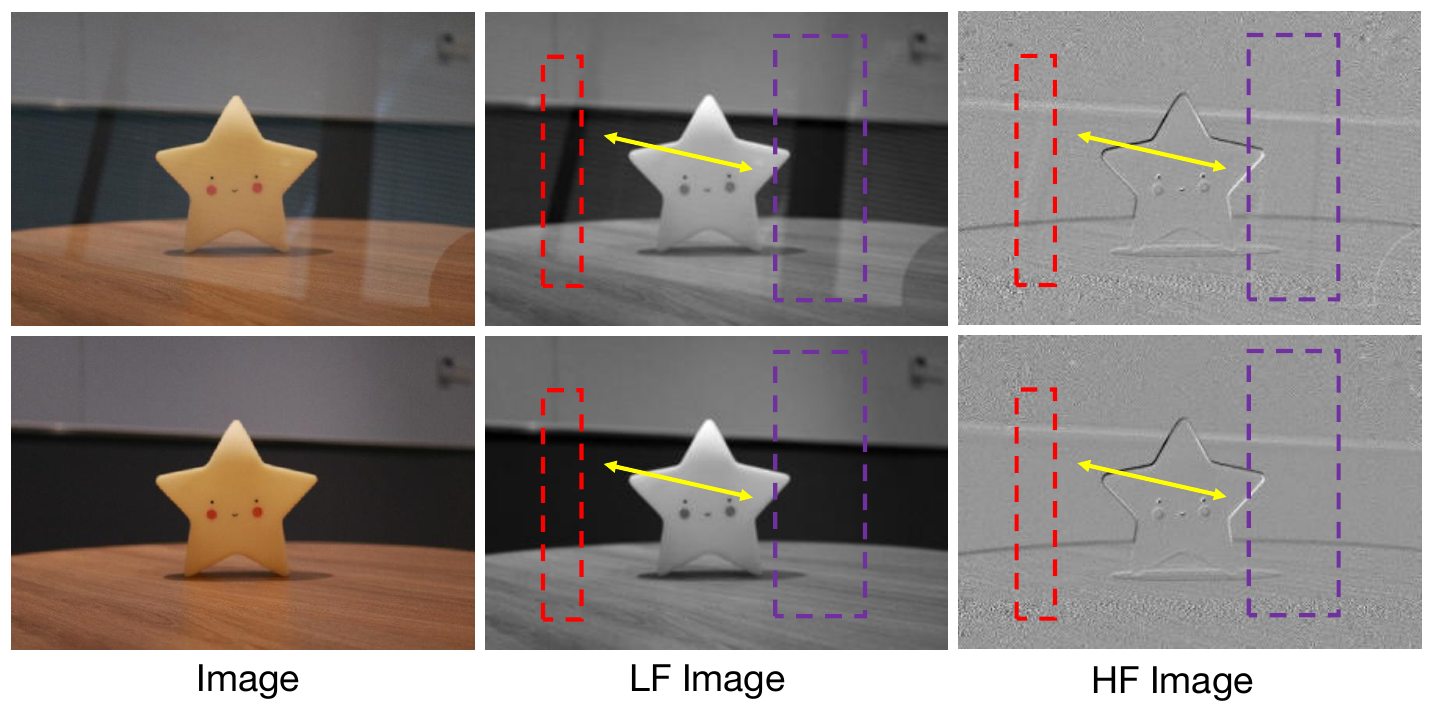}
	  % \vspace{-0.1in}
   % \vspace{-3mm}
	\caption{Illustration of the low-frequency and high-frequency images for both reflection and clear images. Both low- and high-frequency images of the reflected image exhibit distinct differences compared to those of the clear image. These differences tell if a region is reflection-dominated. This motivates us to develop a new SIRR method discerning and employing clear low and high frequencies as prompts to guide the deep model toward better reflection removal.}
	\label{fig:overall_frequency}
 \end{center}
   % \vspace{-4mm}
\end{figure}

Early SIRR methods mainly rely on image priors like gradient sparsity, smoothness, and Gaussian mixture models, but they struggle with complex reflection scenarios. Recent advancements use deep neural networks trained on labeled data to estimate reflection and background components, employing architectures such as CNNs, GANs, RNNs, and transformers for enhanced reflection removal performance. In the case of reflection removal, the presence of reflections is primarily associated with the LF (low frequency) and HF (high frequency) components of images~\cite{chung2009interference,wan2018region}. 
As shown in Fig.~\ref{fig:overall_frequency}, both the reflected scene's low- and high-frequency images show significant deviations compared to those of the corresponding clear image counterparts. These differences tell if a region is reflection-dominated, providing valuable prompts for enhancing the network's performance in removing reflections.
However, it is rarely investigated in deep learning-based SIRR network design.

In this paper, we exploit the LF and HF information as a new visual prompt to guide the deep model for better reflection removal. To achieve this goal, there are mainly two important aspects in the model design: \textit{(1) how to generate accurate LF and HF prompts}, and \textit{(2) how to effectively utilize these prompts in the model to enhance reflection removal}. Some methods~\cite{Liu_2018_CVPR_Workshops,gao2024exploring} consider low and high frequencies of low-quality images in the model, which can improve the restoration performance. However, these methods are not specifically designed for SIRR and do not fully use the frequency information to help enhance image restoration. This limitation leads to their sub-optimal performance.
To generate accurate LF and HF prompts and use these prompts for effective reflection removal, we decouple the reflection removal process into the prompt generation and subsequent prompt-guided restoration. For the prompt generation, we first propose a prompt pre-training strategy to train a frequency prompt encoder that can encode the ground-truth image into LF and HF prompts. Then, we train diffusion models (DMs) to generate LF and HF prompts estimated by the pre-trained frequency prompt encoder. Benefiting from the strong generation ability of DMs, we obtain accurate LF and HF prompts. For the prompt-guided restoration, we integrate the generated prompts into a tailored prompt-based transformer network that utilizes these prompts for effective reflection removal.

In particular, we propose a novel prompt-guided reflection removal method (PromptRR), which effectively utilizes frequency prompts to enhance the model's capability to restore clear images. PromptRR consists of a frequency prompt encoder (FPE), two diffusion models (DMs), and a prompt transformer reflection network (PromptFormer). The process of PromptRR includes prompt generation and prompt-guided restoration. For the prompt generation, PromptRR first adopts the strategy of prompt pre-training with FPE and PromptFormer, which aims to train the FPE so that it can encode accurate frequency prompts from the input ground truth image. Then, DMs are trained to generate the prompts encoded by the per-trained FPE. Finally, PromptRR utilizes the prompts generated by DMs to guide the PromptFormer during the reflection removal process. In PromptFormer, the key component is the proposed Transformer-based prompt block (TPB), which contains
a prompt multi-head self-attention (PMSA) and a prompt feed-forward network (PFFN). In PMSA and PFFN, the core component is a prompt interaction and injection module. This module adaptively injects prompts into deep features, which guides the network to pay more attention to reflections for efficient reflection removal. Extensive evaluations against state-of-the-art methods on public real-world datasets demonstrate the superiority of PromptRR for SIRR.

This paper's contributions are fourfold: (1) To the best of our knowledge, we make the first attempt to utilize frequency information as a new visual prompt in deep models to address the SIRR problem.
(2) A novel SIRR framework PromptRR adopts diffusion models as prompt generators to generate high-quality low- and high-frequency prompts and uses these prompts for effective reflection removal. (3) A tailored PromptFormer is built by the proposed transformer-based prompt blocks, which help the model efficiently use frequency prompts for better reflection removal. (4) Extensive experiments on public real-world datasets show that PromptRR outperforms state-of-the-art methods.

\section{Related Work}
\label{sec:related_work}

\noindent \textbf{Single Image Reflection Removal}. When capturing an image through transparent surfaces, reflection is an undesirable phenomenon. The reflection degradation can be modeled as~\cite{schechner2000separation,Wei_2019_CVPR}:
\begin{equation}
\begin{aligned}
\mathbf{Q}=\mathbf{B}+\mathbf{R} * \mathbf{K},
\end{aligned}
\end{equation}
where $\mathbf{Q}$ denotes the reflection-contaminated image, $\mathbf{B}$ refers to the desired background layer, $\mathbf{R}$ is the reflection content, $*$ represents a convolution operator, and $\mathbf{K}$ is a Gaussian blur kernel. The above degradation model shows that SIRR is an ill-posed problem. Prior-based methods aim to extract the desired prior information from the reflection and background layers to help the model effectively achieve reflection removal. For example, sparsity prior~\cite{levin2007user}, gradient sparsity prior~\cite{levin2004separating}, and Gaussian mixture model prior~\cite{shih2015reflection} are used for reflection removal. Unfortunately, these priors usually rely on certain physical models that do not fully encapsulate complex scenes, and thus the performance of prior-based methods degrades. 

Recently, numerous deep learning-based techniques~\cite{zhang2018adversarial,chang2018single,zhang2021deep,zhang2021single,li2020improved,hong2024light} have been proposed and successfully applied in the design of SIRR methods. For example, Zhang \etal \cite{zhang2018single} incorporate perceptual information into a conditional generative adversarial network (GAN). Wen \etal (WY19)~\cite{Wen_2019_CVPR} propose an alignment-invariant loss that benefits training on unaligned real-world images. In \cite{Kim_2020_cvpr}, Kim \etal consider the spatial variability of reflections' visual effects for reflection removal. On the other hand, Song \etal \cite{song2023robust} propose RSIRR, a robust SIRR network using Transformers, while multi-stage SIRR methods refine restored results progressively through network design. For instance, some methods employ two-stage (ZS21~\cite{Zheng_2021_CVPR}) or three-stage (YG18~\cite{Yang_2018_ECCV} and CL21~\cite{Chang2021_wacv}) deep networks for reflection removal. Additionally, Li \etal (LY20)~\cite{Li_2020_CVPR} propose a multi-stage network that leverages long cascade networks with Long Short-Term Memory to achieve reflection removal. In~\cite{hu2021trash}, Hu and Guo (HG21) incorporate interactions between two network branches, which can utilize information more effectively. Recently, DSRNet~\cite{hu2023single} employs a mutually gated interaction mechanism using a two-stage structural design for SIRR. Zhu \etal \cite{zhu2024revisiting} propose a maximum reflection filter for estimating reflection locations in SIRR. 

Beyond these, Zhao \etal \cite{zhao2025reversible} present a Reversible Decoupling Network that adopts a multi-column reversible encoder and a transmission-rate-aware prompt generator to preserve semantic information while efficiently disentangling reflection and transmission layers. Furthermore, Huang \etal \cite{huang2025lightweight} propose a lightweight Deep Exclusion Unfolding Network, which integrates the deep unfolding paradigm with exclusion priors to achieve computational efficiency without sacrificing performance. In addition, He \etal \cite{he2025rethinking} revisit the role of depth guidance and propose a depth-aware reflection removal framework, which adaptively leverages depth information to enhance separation between layers. Chen \etal \cite{chen2024real} address the challenge of Ultra-High-Definition image reflection removal and propose a U-Net-based architecture tailored for single-image reflection removal, specifically optimized for the complexities of high-resolution image processing. However, existing methods still overlook critical low- and high-frequency details. Our proposed PromptRR addresses this limitation by leveraging frequency prompts and a Transformer-based network for more accurate reflection removal.

% \vspace{1mm}
\noindent \textbf{Diffusion Models}. Diffusion probabilistic models (DMs) have excelled in tasks like image synthesis and density estimation, offering advantages over other generative models. As a result, more researchers have adopted DMs to tackle challenges in low-level vision, such as image super-resolution~\cite{SRDiff2022, Super_TPAMI2022,wang2025rap,chen2025unsupervised}, image inpainting~\cite{Lugmayr_2022_CVPR}, image enhancement~\cite{yan2025efficient}, and image restoration~\cite{Palette2022, Rombach_2022_CVPR,zhang2025ssp,cui2024omni}. For example, the latent diffusion model~\cite{Rombach_2022_CVPR} is proposed to enhance restoration efficiency by implementing a diffusion process in the latent space. RePaint~\cite{Lugmayr_2022_CVPR} improves the denoising strategy to achieve higher visual quality for image inpainting. Palette~\cite{Palette2022}, through conditional diffusion models, advances the state of the art in four demanding image-to-image translation tasks. These methods primarily concentrate on generating new content for individual pixels and exhibit high fidelity in their output generation. In contrast, our method conducts a diffusion process to predict frequency prompts to help the model achieve more effective reflection removal.

\vspace{1mm}
\noindent \textbf{Prompt Learning}. Prompt-based learning has recently garnered significant attention in natural language processing (NLP) and computer vision. Prompt-based methods involve conditioning pre-trained models with additional instructions to achieve specific tasks~\cite{brown2020language}, and prompt plays a key role in downstream datasets~\cite{Zhou2022COOp}. 
Since using specific manual instruction sets as prompts requires domain expertise and is time-consuming~\cite{Zhou2022COOp}, some recent studies have started discussing the possibility of applying learnable prompts to achieve better performance on vision tasks~\cite{Zhou2022COOp, VPT_2022, Khattak_2023_CVPR}, incremental learning~\cite{wang2022dualprompt, Wang_2022_CVPR}, and multitask learning~\cite{he2022hyperprompt,wang2023multitask}. However, most existing studies focus on high-level vision problems, and the potential of using prompt learning for SIRR remains unexplored. In our work, we propose a novel approach that combines prompt learning and diffusion models for SIRR. By using learnable prompts, we help models better remove reflections.

\section{Methodology}
\label{sec:method}

\begin{figure}[t]
\begin{center}
 % \vspace{-4mm}
	\includegraphics[width=0.45\textwidth]{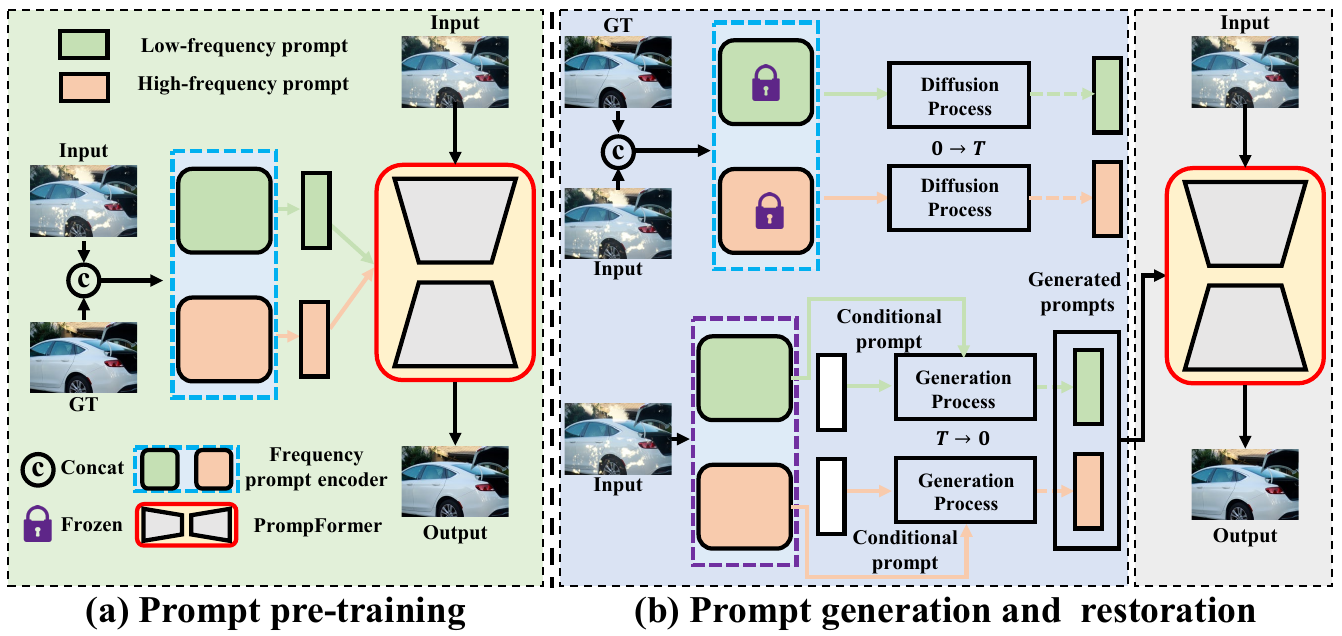}
	  % \vspace{-0.1in}
   % \vspace{-3mm}
	\caption{An overview of our PromptRR. It includes two main stages: (a) prompt pre-training stage and (b) prompt generation and restoration stage. Notably, we do not use the ground-truth image (GT) in the inference stage.}
	\label{fig:overall_framework}
 \end{center}
   % \vspace{-4mm}
\end{figure}

% \subsection{Overall Pipeline}
In this paper, we propose PromptRR, a frequency prompt learning-based framework for removing reflections. PromptRR effectively utilizes learned frequency prompts to enhance the model's ability to recover clear images. As shown in Fig.~\ref{fig:overall_framework}, PromptRR first adopts the strategy of prompt pre-training with a prompt encoder and a reflection removal network PromptFormer. The goal of the prompt pre-training is to train the prompt encoder to extract the LF and HF prompts from input images accurately. Then, the trained prompt encoder is utilized in the diffusion models to generate LF and HF prompts from the input degraded image. Finally, PromptRR utilizes these prompts to guide PromptFormer during the reflection removal process. Next, we provide details of our method's prompt pre-training and prompt generation and recovery phases.

\subsection{Prompt Pre-training}
Recent advances in NLP-based text prompts have shown their potential in guiding models and improving prediction accuracy~\cite{liu2023pre}. However, applying text prompts directly to vision tasks, such as single reflection removal, poses challenges due to the absence of accompanying text. Although previous methods~\cite{herzig2022promptonomyvit,gan2023decorate,wang2023selfpromer} have explored text-free prompts like depth and segmentation prompts for vision tasks, these prompts may not be suitable for reflection removal. Therefore, our method introduces frequency cues as a new visual cue to guide models to remove reflections more effectively. To achieve this goal, we propose a frequency prompt encoder (FPE) and a reflection removal network (PromptFormer) to conduct the prompt pre-training for the subsequent prompt generation by diffusion models. Next, we discuss the structures of FPE and PromptFormer and present the details of the prompt pre-training. 

\begin{figure*}[t]
\begin{center}
   \includegraphics[width=\textwidth]{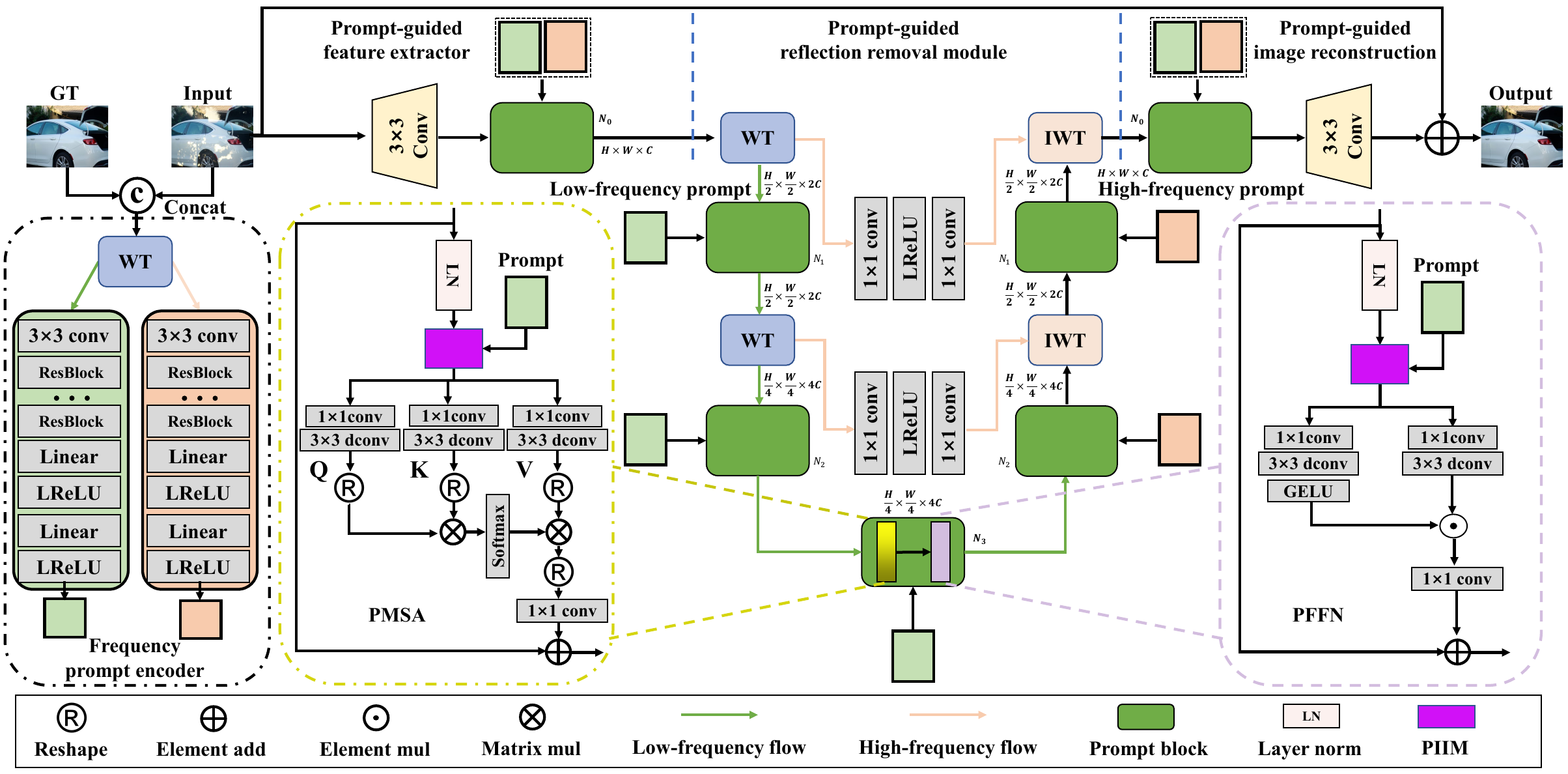}
	  % \vspace{-0.1in}
   % \vspace{-2mm}
	\caption{The overview of FPE and PromptFormer for the prompt pre-training. FPE includes a wavelet transform and a dual-branch encoder. PromptFormer is mainly built by a transformer-based prompt block consisting of the prompt multi-head self-attention (PMSA) and the prompt feed-forward network (PFFN).  WT and IWT are the wavelet transform and inverse wavelet transform respectively.}	\label{fig:overall_promptformer}
 \end{center}
   % \vspace{-6mm}
\end{figure*}

As shown in Fig.~\ref{fig:overall_promptformer} (left), FPE mainly includes a wavelet transform~\cite{huang2021selective,zhou2023xnet} and a dual-branch encoder. Within the dual-branch encoder, each branch mainly comprises residual blocks and linear layers.
In FPE, the input images are fed to wavelet transform, resulting in LF and HF images. These images are then fed into the LF branch (depicted in green in Fig.~\ref{fig:overall_promptformer}) and the HF branch (shown in light orange in Fig.~\ref{fig:overall_promptformer}) to generate LF and HF prompts, respectively. After that, PromptFormer adopts the generated prompts as guidance for better reflection removal. The overall structure of PromptFormer is shown in Fig.~\ref{fig:overall_promptformer} (right). PromptFormer is a transformer-based reflection removal network that includes a prompt-guided feature extractor, a prompt-guided reflection removal module, and a prompt-guided image reconstruction. The core block in PromptFormer is the transformer-based prompt block (TPB). We stack $N_{i \in[0,1,2,3]}$ TPBs to build the prompt-guided reflection removal module. 

% \begin{figure*}[t]
% \begin{center}
%  % \vspace{-4mm}
%   \includegraphics[width=0.8\textwidth]{images/PIPM.pdf}
% 	  % \vspace{-0.1in}
%   \vspace{-3mm}
% 	\caption{Architecture of the prompt interaction and injection module (PIIM). PIIM contains two stages: Prompt interaction and injection. In the feature extractor and image reconstruction stages of PromptFormer, we only use prompt injection in TPBs.}
% 	\label{fig:PIIM}
%  \end{center}
%    % \vspace{-6mm}
% \end{figure*}

\begin{figure}[t]
\begin{center}
 % \vspace{-4mm}
  \includegraphics[width=0.5\textwidth]{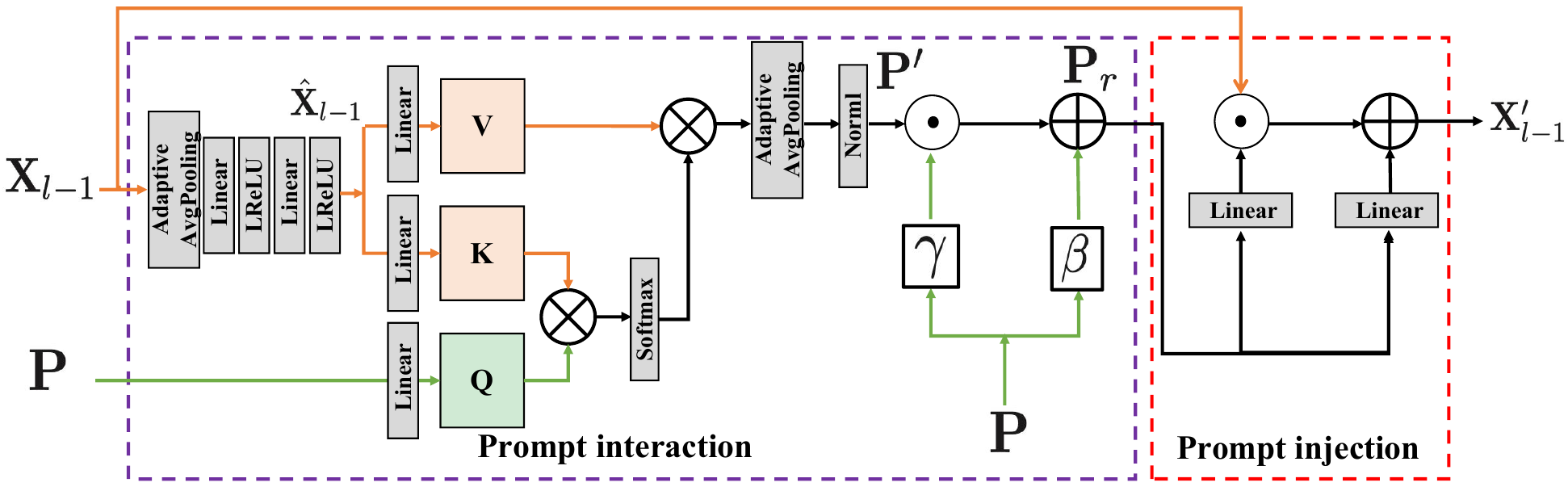}
	  % \vspace{-0.1in}
  % \vspace{-3mm}
	\caption{Structure of the prompt interaction and injection module. It has two stages: Prompt interaction and injection. In the feature extractor and image reconstruction stages of PromptFormer, we only use prompt injection in TPBs.}
	\label{fig:PIIM}
 \end{center}
   % \vspace{-6mm}
\end{figure}

% \begin{figure}[ht]
% \begin{center}
%     \subfloat[Architecture of PIIM]{\includegraphics[width=0.45\textwidth]{images/figure4-a.pdf}}
%     \quad
%     \subfloat[Architecture of PIIM]
%     {\includegraphics[width=0.4\textwidth]{images/figure4-b.pdf}}
%     \caption{\textbf{(a)} Architecture of the prompt interaction and injection module (PIIM). PIIM contains two stages: Prompt interaction and injection. In the feature extractor and image reconstruction stages of PromptFormer, we only use prompt injection in TPBs. \textbf{(b)} The overall structure of the prompt generation where we use the low-frequency generation diffusion model as an example.}\label{fig:figure1}
%     \vspace{-0.2in}
% \end{center}
% \end{figure}

The transformer-based prompt block contains a prompt multi-head self-attention (PMSA) and a prompt feed-forward network (PFFN), ensuring the Transformers for better reflection removal using the learned frequency prompts. The proposed TPB can be formulated as,
% \vspace{-2mm}
\begin{equation}
\begin{aligned}
&\mathbf{X}_l^{\prime}=\mathbf{X}_{l-1}+\text{PMSA}\left(\text{LN}\left(\mathbf{X}_{l-1}\right), \mathbf{P}\right),\\
&\mathbf{X}_l=\mathbf{X}_l^{\prime}+\text{PSFN}\left(\text{LN}\left(\mathbf{X}_l^{\prime}\right), \mathbf{P}\right),
\end{aligned}
% \vspace{-2mm}
\end{equation}
where $\mathbf{X}_{l-1}$ and $\mathbf{X}_{l}$ refer to the input and output features of the $l^{\mathrm{th}}$ TPB, and $\mathbf{P}$ is LF or HF prompt. 

\vspace{1mm}
\noindent \textbf{Prompt Multi-head Self-attention}. Existing Transformers~\cite{zamir2022restormer,wang2022uformer,song2023robust} typically extract query ($Q$), key ($K$), and value ($V$) from the input feature to perform multi-head self-attention. However, these Transformers may not effectively represent features without explicit visual cues. To mitigate this challenge, we develop PMSA in the Transformers, aiming to enhance the network's focus on extra visual prompts and thus improve its performance for reflection removal. Specifically, we propose a prompt interaction and injection module (PIIM), refining the input feature using prompts before generating $Q$, $K$, and $V$. The structure of PIIM is shown in Fig.~\ref{fig:PIIM}. It consists of two steps: prompt interaction and prompt injection. Specifically, for the prompt interaction step, we first use adaptive average pooling followed by two linear layers to transform input feature $\mathbf{X}_{l-1}$ into $\hat{\mathbf{X}}_{l-1}$ with the same shape with the prompt $\mathbf{P}$. Then, we conduct a cross-attention interaction between $\mathbf{P}$ and $\hat{\mathbf{X}}_{l-1}$ in the spatial dimension. The cross-attention interaction process is shown as,
% \vspace{-2mm}
\begin{equation}
\begin{aligned}
& Q=W^Q\hat{\mathbf{X}}_{l-1}, K_p= W^K \mathbf{P}, V=W^V \hat{\mathbf{X}}_{l-1}, \\
& \mathbf{P}^{\prime}=\operatorname{Adap}(\operatorname{Softmax}\left(\frac{QK^{\top}}{\alpha}\right)V), 
\end{aligned}
% \vspace{-2mm}
\end{equation}
where $W^Q$, $W^K$, and $W^V$ denote the projection matrices of the query, key, and value. $\alpha$ is an optional temperature factor and $\operatorname{Adap}$ is the adaptive average pooling operation. Then, $\mathbf{P}^{\prime}$ is further refined by $\mathbf{P}$ using two learnable fusion parameters~\cite{wei2022hairclip,huang2017arbitrary} shown as,
% \vspace{-2mm}
\begin{equation}
\begin{aligned}
\mathbf{P}_{r} =\frac{\mathbf{P}^{\prime}-\mu}{\sigma}\odot(1+\gamma)+\beta,
\end{aligned}
% \vspace{-2mm}
\end{equation}
where $\mathbf{P}_{r}$ is the refined prompt, $\odot$ is element-wise multiplication, $\mu$ and $\sigma$ are the mean and standard deviation of $\mathbf{P}^{\prime}$ respectively. 
In this formulation, $\gamma$ and $\beta$ are generated by $f_\gamma(P)$ and $f_\beta(P)$ implemented with two simple linear layers with one intermediate layer norm and leaky relu layer. 

For the prompt injection step, the refined prompt $\mathbf{P}_{r}$ is injected into $\mathbf{X}_{l-1}$ to derive the prompt modulation feature $\mathbf{X}_{l-1}^{\prime}$ as,
% \vspace{-2mm}
\begin{equation}
\begin{aligned}
\mathbf{X}_{l-1}^{\prime}=W_{1} \mathbf{P}_{r} \odot \mathbf{X}_{l-1}+ W_{2}\mathbf{P}_{r},
\end{aligned}
% \vspace{-2mm}
\end{equation}
where $W_{1}$ and $W_{2}$ denote the liner layer. After that, we apply a $1\times1$ convolution, followed by a $3\times3$ depth-wise convolution, to generate the query, key, and value from the modulation feature, respectively. Additionally, we perform self-attention across channels instead of the spatial dimension, aiming to reduce the time and memory complexity. This design choice is motivated by recent works~\cite{zamir2022restormer,wang2023ultra,xia2023diffir}.

\vspace{1mm}
\noindent \textbf{Prompt Feed-forward Network}. To make full use of the frequency prompt information in the network, we also introduce our plug-and-play PIIM into the recent feed-forward network~\cite{zamir2022restormer,xia2023diffir} to build our PFFN, which is shown in Fig.~\ref{fig:overall_promptformer} (right). More details are provided in the supplemental material.

For prompt pre-training, we train FPE and PromptFormer together using the $\mathcal{L}_{1}$ loss,
% \vspace{-2mm}
\begin{equation}
\begin{aligned}
\mathcal{L}_{1}=\left\|I_{\prime}-I_{gt}\right\|_1,
\end{aligned}
% \vspace{-2mm}
\end{equation}
where $I_{\prime}$ is the restored result of PromptFormer and $I_{gt}$ is the ground-truth image. With this prompt pre-training, FPE can produce LF and HF prompts accurately for subsequent prompt generation by diffusion models.

% \begin{figure*}[t]
% \begin{center}
%  % \vspace{-4mm}
% 	\includegraphics[width=0.8\textwidth]{images/diffusion_models.pdf}
%    % \vspace{-3mm}
% 	\caption{The overall structure of the prompt generation where we use the low-frequency generation diffusion model as an example.}
% 	\label{fig:diffusion}
%  \end{center}
%    % \vspace{-5mm}
% \end{figure*}

\begin{figure}[t]
\begin{center}
 % \vspace{-4mm}
	\includegraphics[width=0.5\textwidth]{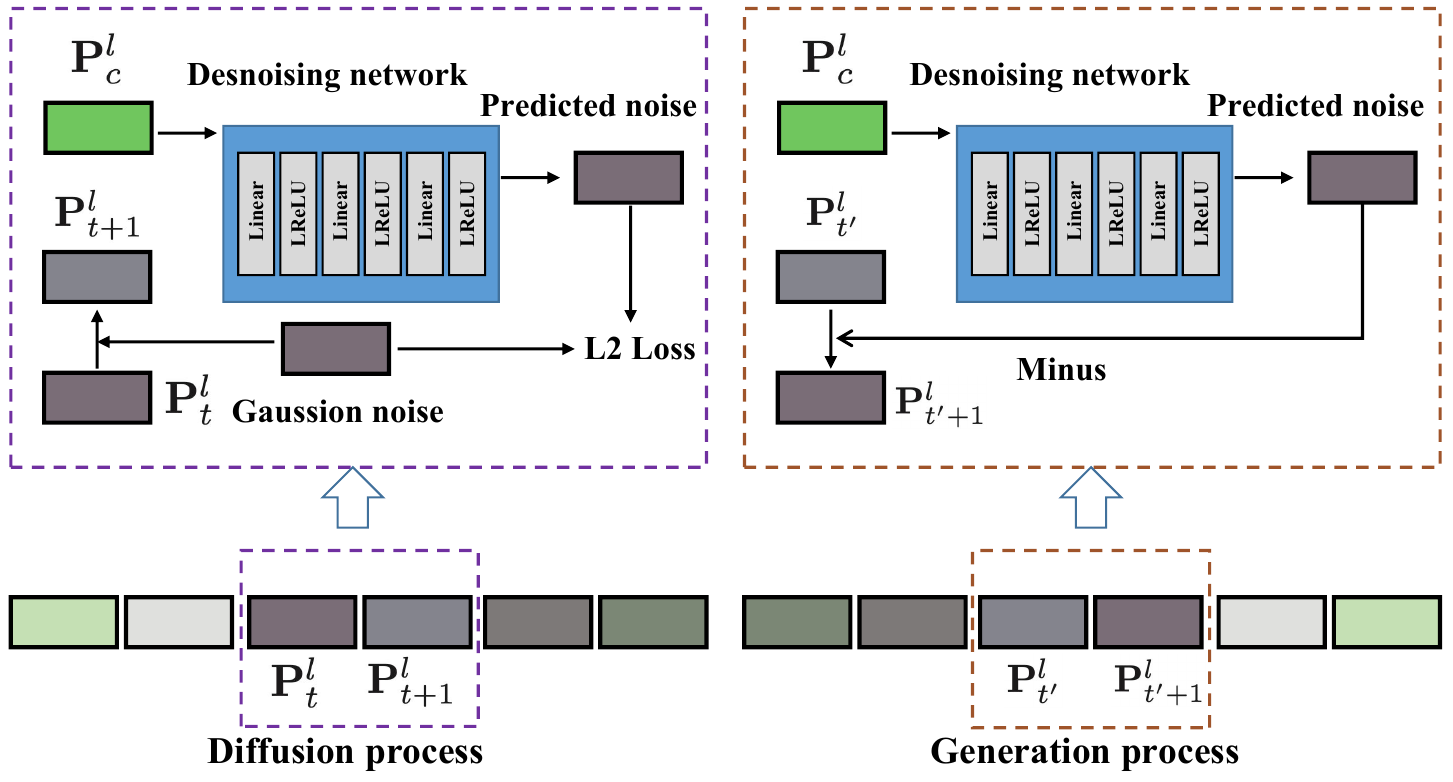}
   % \vspace{-3mm}
	\caption{The overall structure of the prompt generation, where we use the low-frequency generation diffusion model as an example.}
 	% \caption{The overall architecture of the prompt generation by the diffusion model where we use low-frequency generation diffusion as an example.}
	\label{fig:diffusion}
 \end{center}
   % \vspace{-5mm}
\end{figure}

\begin{algorithm}[t]
\hspace*{\algorithmicindent}\noindent \textbf{Input:} reflection image $I$, ground-truth image $I_{gt}$, pre-trained frequency prompt encoder $\operatorname{FPE_{pre}}$, and conditional frequency prompt encoder $\operatorname{FPE_{con}}$.
\caption{Diffusion training (diffusion process)}\label{algo_train}
\begin{algorithmic}[1]
        \While{not converged}
        \State $\mathbf{P}^{l}_0,\mathbf{P}^{h}_0 = \operatorname{FPE_{pre}}(\operatorname{Concat}(I_{gt},I)). $
        \State $\mathbf{P}^{l}_c,\mathbf{P}^{h}_c = \operatorname{FPE_{con}}(I). $
        \State $t \sim \text{Uniform}\{1, \ldots, T\}$
        \State $\bm{\epsilon^{l}} \sim \mathcal{N}(\mathbf{0},\mathbf{I})$
        \State $\bm{\epsilon^{h}} \sim \mathcal{N}(\mathbf{0},\mathbf{I})$
        \State $\mathbf{P}^{l}_{t+1} = \sqrt{\bar{\alpha}^{l}_t} \mathbf{P}^{l}_0+\sqrt{1-\bar{\alpha}^{l}_t} \bm{\epsilon^{l}}$
        \State $\mathbf{P}^{h}_{t+1} = \sqrt{\bar{\alpha}^{h}_t} \mathbf{P}^{h}_0+\sqrt{1-\bar{\alpha}^{h}_t} \bm{\epsilon^{h}}$
        \State $\mathbf{e}^{l}_t = \bm{\epsilon^{l}}_\theta\left(\mathbf{P}^{l}_c, \mathbf{P}^{l}_{t+1}, t\right)$
        \State $\mathbf{e}^{h}_t = \bm{\epsilon^{h}}_\theta\left(\mathbf{P}^{h}_c, \mathbf{P}^{h}_{t+1}, t\right)$  
      \State Perform gradient descent steps on $\nabla_\theta 
        \;\mathcal{L}_{\mathrm{diff}}(\theta)$ 
     % \EndIf
        \EndWhile
      \\ \textbf{Output:} the trained low-frequency prompt generation diffusion model $\bm{\epsilon^{l}}_\theta$, the trained high-frequency prompt generation diffusion model $\bm{\epsilon^{h}}_\theta$,  the trained conditional frequency prompt encoder $\operatorname{FPE_{con}}$.    
      % \\ $\textbf{return}\;\; \theta$ 
\end{algorithmic}
  % \vspace{-2mm}
\end{algorithm}

\subsection{Prompt Generation and Restoration}
To generate more accurate prompts for better reflection removal, we present a dual-diffusion model as a prompt generator by taking advantage of the powerful generative abilities of diffusion models. As shown in Fig.~\ref{fig:overall_framework}, our prompt generator consists of a low-frequency generation diffusion model and a high-frequency generation diffusion model. These two diffusion models share the same architecture, where the denoise network is implemented by several linear layers with leaky relu~\cite{huang2017arbitrary,xia2023diffir}. The diffusion model contains two processes, \ie the diffusion process and the generation process. For ease of illustration, we use the low-frequency generation diffusion as an example of these two processes in the following.

% \input{table_latex/table1_comparison_results}
% \subsubsection{Prompt Generator}
\vspace{1mm}
\noindent \textbf{Diffusion Process}. During the diffusion process, we first adopt the pre-trained FPE to produce initial LF and HF prompt vectors (\ie $\mathbf{P}^{l}$ and $\mathbf{P}^{h}$). Then, continuously adding Gaussian noise to the initial prompt vectors, the characteristics of the prompt vectors will gradually disappear, and the
prompt eventually becomes standard Gaussian noise. 
Specifically, as illustrated in Fig.~\ref{fig:diffusion}, at the diffusion step $t$, the next noisy prompt vector $\mathbf{P}^{l}_{t+1}$ can be obtained by $\mathbf{P}^{l}_{t+1}=\mathbf{P}^{l}_{t}+\epsilon$, where $\epsilon$ is the Gaussian noise. After that, the next noisy prompt and the proposed conditions are fed into the denoise network $\epsilon_\theta$ to estimate the distribution of the noise, which is optimized by the squared error loss,
% \vspace{-2mm}
\begin{equation}
\begin{aligned}
\mathcal{L}^{l}_{diff}=\left\|\epsilon-\epsilon_\theta\left(\mathbf{P}^{l}_{c}, \mathbf{P}^{l}_{t+1},t\right)\right\|^2,
\end{aligned}
% \vspace{-2mm}
\end{equation}
where $\mathbf{P}^{l}_{c}$ represents the conditional prompt that is extracted from the reflection-contaminated input by FPE. Specifically, given the diffusion step, $\mathcal{L}^{l}_{diff}$ can be directly calculated by the original prompt rather than the intermediate status. The detail of this process is shown in Algorithm~\ref{algo_train}.

\begin{algorithm}[t]
\hspace*{\algorithmicindent}\noindent \textbf{Input:} reflection image $I$, the number of implicit sampling steps $T$, the trained low-frequency prompt generation diffusion model $\bm{\epsilon^{l}}_\theta$, the trained high-frequency prompt generation diffusion model $\bm{\epsilon^{h}}_\theta$, and the trained conditional frequency prompt encoder $\operatorname{FPE_{con}}$.
\caption{Diffusive sampling (generation process)}\label{algo}
\begin{algorithmic}[1]
        \State $\mathbf{P}^{l}_c,\mathbf{P}^{h}_c = \operatorname{FPE_{con}}(I). $
      \State $\mathbf{P}^{l}_T \sim \mathcal{N}(\mathbf{0}, \mathbf{I})$
     \State $\mathbf{P}^{h}_T \sim \mathcal{N}(\mathbf{0}, \mathbf{I})$
      \For{$t =T, \ldots, 1$}
        \State $\mathbf{e}^{l}_{t-1} = \bm{\epsilon^{l}}_\theta\left(\mathbf{P}^{l}_c, \mathbf{P}^{l}_t,  t \right)$
        \State $\mathbf{P}^{l}_{t-1} \!\!= \!\!\sqrt{\bar{\alpha}^{l}_{t-{1}}}\left(\frac{\mathbf{P}^{l}_{t}-\sqrt{1-\bar{\alpha}^{l}_t} \cdot \mathbf{e}^{l}_{t-1}}{\sqrt{\bar{\alpha}^{l}_t}}\right)
       \! +\!\sqrt{1-\bar{\alpha}^{l}_{t-{1}}} \cdot \mathbf{e}^{l}_{t-1}$

        \State $\mathbf{e}^{h}_{t-1} = \bm{\epsilon^{h}}_\theta\left(\mathbf{P}^{h}_c, \mathbf{P}^{h}_t,  t \right)$
        \State $\mathbf{P}^{h}_{t-1} \!\!= \!\!\sqrt{\bar{\alpha}^{h}_{t-{1}}}\left(\frac{\mathbf{P}^{h}_{t}-\sqrt{1-\bar{\alpha}^{h}_t} \cdot \mathbf{e}^{h}_{t-1}}{\sqrt{\bar{\alpha}^{h}_t}}\right)
       \! +\!\sqrt{1-\bar{\alpha}^{h}_{t-{1}}} \cdot \mathbf{e}^{h}_{t-1}$
      \EndFor
      % \State \Return $\mathbf{P}^{l}_0$, $\mathbf{P}^{h}_0$

     \\ \textbf{Output:} low-frequency prompt $\mathbf{P}^{l}_0$, high-frequency prompt $\mathbf{P}^{h}_0$. 
\end{algorithmic}
  % \vspace{-1mm}
\end{algorithm}
\vspace{1mm}
\noindent \textbf{Generation Process}. In the generation process, the prompt vector is finally generated by gradually denoising the random initial Gaussian noise vector under the conditional prompt from the FPE. Specifically, at the generation step $t^{\prime}$, the denoising prompt vector $\mathbf{P}^{l}_{t^{\prime}}$ and the conditional prompt $\mathbf{P}^{l}_{c}$ are fed into the denoise network to predict the noise. 

Then, the next denoising prompt vector $\mathbf{P}^{l}_{t^{\prime}+1}$ is obtained by subtracting the predicted noise from the current denoising prompt vector. This process can be formulated as,
% \vspace{-2mm}
\begin{equation}
\begin{aligned}
\mathbf{P}^{l}_{t^{\prime}+1}=\mathbf{P}^{l}_{t^{\prime}}-\epsilon_\theta\left(\mathbf{P}^{l}_{c}, \mathbf{P}^{l}_{t^{\prime}},t^{\prime}\right).
\vspace{-1mm}
\end{aligned}
\end{equation}
The detail of this process is illustrated in 
Algorithm~\ref{algo}. After the dual-diffusion model generates the frequency prompts $\hat{\mathbf{P}}^{l}_{\textit{T}\rightarrow0}$ and $\hat{\mathbf{P}}^{h}_{\textit{T}\rightarrow0}$, our PromptFormer incorporates them as guidance for better reflection removal. Specifically, in the prompt generation and restoration stage, we first train this diffusion model individually in some iterations and then train PromptFormer and the diffusion model together. The joint training loss $\mathcal{L}$ is,
% \vspace{-2mm}
\begin{equation}
\begin{aligned}
\mathcal{L}=\mathcal{L}^{l}_{diff} + \mathcal{L}^{h}_{diff}+ \mathcal{L}_{1},
\end{aligned}
% \vspace{-2mm}
\end{equation}
where $\mathcal{L}^{l}_{diff}$ and $\mathcal{L}^{h}_{diff}$ denote the diffusion losses of low-frequency diffusion and the high-frequency diffusion models. $\mathcal{L}_{1}$ is the pixel-wise loss from PromptFormer. 

For inference, given an input image, our model first generates LF and HF prompts using a diffusion model. The resulting prompts and input images are fed into PromptFormer to restore the final clear image.

%%%% comparison results
\begin{table*}[t]\small
   \centering
   % \vspace{-3mm}
       \caption{Comparison of quantitative results on commonly used real-world datasets in terms of PSNR and SSIM. \textbf{Bold} and \underline{underline} indicate the best and second-best results.}
   % \resizebox{1\textwidth}{!}{
    \begin{tabular}{cc|cc|cc|cc|cc|cc|cc}
    \hline 
    \multicolumn{2}{c|}{\multirow{2}{*}{Methods}}
    & \multicolumn{2}{c|}{\multirow{2}{*}{\textit{Nature}~\cite{Li_2020_CVPR}}}  &  \multicolumn{8}{c|}{\textit{$SIR^{2}$}~\cite{Wan_2017_iccv}}  & \multicolumn{2}{c}{\multirow{2}{*}{\textit{Real}~\cite{zhang2018single}}}\\ \cline{5-12}
        & &  &  &  \multicolumn{2}{c|}{\textit{Postcard}} &  \multicolumn{2}{c|}{\textit{SolidObject}} &  \multicolumn{2}{c|}{\textit{WildScene}} & \multicolumn{2}{c|}{\textit{Average}} & \\ \hline
    & & PSNR & SSIM & PSNR & SSIM & PSNR   & SSIM  & PSNR   & SSIM  & PSNR   & SSIM& PSNR   & SSIM \\ \hline

    & WY19~\cite{Wei_2019_CVPR} & 19.54& 0.7390 & 17.02 &0.7528 & 22.28 & 0.8072 &22.07 & 0.8196& 20.27&0.7895 & 21.82 & 0.7599 \\ %\hline

    & WT19~\cite{Wen_2019_CVPR} &10.45 & 0.1593   & 7.63 & 0.2871&10.81& 0.2359 & 11.16& 0.1945 &9.70 & 0.2463& 11.66  &0.0928   \\ %\hline

     & LY20~\cite{Li_2020_CVPR} & 19.41 & 0.7557 & 17.58 & 0.7807 & 23.34& 0.8480 & 23.27&0.8855  & 21.18& 0.8308& 22.65& 0.7785 \\ 

   & ZS21~\cite{Zheng_2021_CVPR} &  18.07& 0.7496 & 16.11  &0.7507  & 19.65 & 0.7538  &20.02 &0.8228
&18.41 &0.7672 & 19.15 & 0.6659  \\ %\hline
     & CL21~\cite{Chang2021_wacv} &  20.02  & 0.7786  & 18.89& 0.7861 &  23.42& 0.8327& 24.02&0.8964 & 21.86& 0.8287&22.12 &0.7721 \\ 
   & HG21~\cite{hu2021trash} & 20.44& \underline{0.7901} & 18.65& 0.8076 & 24.05 & 0.8566 &23.86 &0.8897 & 22.00& 0.8453& 22.11 &0.7751 \\ 

   & Uformer~\cite{wang2022uformer} & 19.94 &0.7826 &19.28& 0.7822& 22.61& 0.8226& 22.91& 0.8923 & 21.43& 0.8222& 20.04 & 0.7267 \\ 
   
  &Restormer~\cite{zamir2022restormer} & 20.71 &0.7794 & 17.80& 0.7848& 24.07 & 0.8570 & 24.04&0.8490 & 21.73&0.8388 & 20.38 & 0.7315 \\

   & RSIRR~\cite{song2023robust} & \underline{20.97} &0.7864  & \underline{19.65}  & \underline{0.8230} & \textbf{24.71}& \textbf{0.8700}& \underline{24.70} & \underline{0.8976}& \underline{22.82} &  \underline{0.8583} &\underline{23.61} &\underline{0.7912} \\  \hline
    & \textbf{PromptRR (Ours)} & \textbf{21.00} & \textbf{0.8142} & \textbf{23.03} & \textbf{0.8653} & \underline{24.17} &\underline{0.8587}& \textbf{26.43} & \textbf{0.9300} & \textbf{24.22} & \textbf{0.8761} &\textbf{24.11} & \textbf{0.8126}  \\ \hline
    \end{tabular}
    % }

     \label{tab:quanti_res}
       % \vspace{-1mm}
\end{table*}

\def \rootnaturev2 {images/datasets-v2/Nature/origin_img_with_box/}
\begin{figure*}[ht]
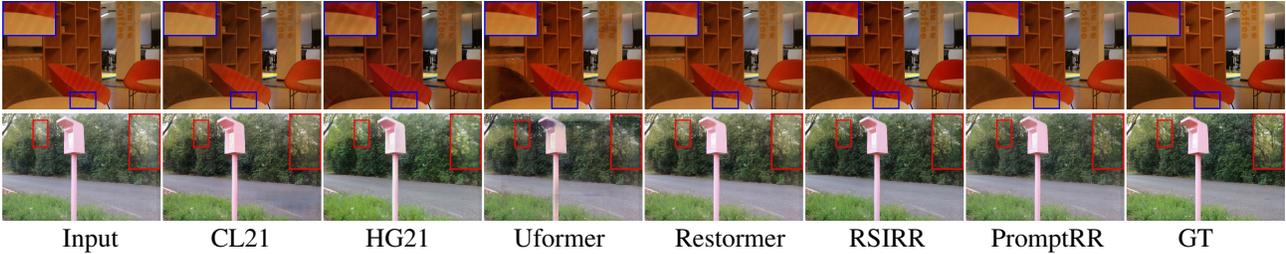

    \centering
    \resizebox{1\textwidth}{!}{
    \begin{minipage}{\textwidth}
      \centering
    \begin{subfigure}[t]{0.24\linewidth}
        \includegraphics[width=\linewidth]{\rootnaturev2 input/input_1-2_71.pdf}
        \caption{\small Input (PSNR: 23.55)}
    \end{subfigure} \hspace{-5pt}
    \begin{subfigure}[t]{0.24\linewidth}
        \includegraphics[width=\linewidth]{\rootnaturev2 WY19/WY19_1-2_71.pdf}
        \caption{\small WY19 (PSNR: 19.37)}
    \end{subfigure} \hspace{-5pt}
    \begin{subfigure}[t]{0.24\linewidth}
        \includegraphics[width=\linewidth]{\rootnaturev2 WT19/WT19_1-2_71.pdf}
        \caption{\small WT19 (PSNR: 10.05)}
    \end{subfigure} \hspace{-5pt}
    \begin{subfigure}[t]{0.24\linewidth}
        \includegraphics[width=\linewidth]{\rootnaturev2 LY20/LY20_1-2_71.pdf}
        \caption{\small LY20 (PSNR: 21.95)}
    \end{subfigure} \\

    \begin{subfigure}[t]{0.24\linewidth}
        \includegraphics[width=\linewidth]{\rootnaturev2 ZS21/ZS21_1-2_71.pdf}
        \caption{\small ZS21 (PSNR: 16.79)}
    \end{subfigure} \hspace{-5pt}
    \begin{subfigure}[t]{0.24\linewidth}
        \includegraphics[width=\linewidth]{\rootnaturev2 CL21/CL21_1-2_71.pdf}
        \caption{\small CL21 (PSNR: 21.94)}
    \end{subfigure} \hspace{-5pt}
    \begin{subfigure}[t]{0.24\linewidth}
        \includegraphics[width=\linewidth]{\rootnaturev2 HG21/HG21_1-2_71.pdf}
        \caption{\small HG21 (PSNR: 19.24)}
    \end{subfigure} \hspace{-5pt}
    \begin{subfigure}[t]{0.24\linewidth}
        \includegraphics[width=\linewidth]{\rootnaturev2 Uformer/Uformer_1-2_71.pdf}
        \caption{Uformer (PSNR: 22.43)}
    \end{subfigure} \\

    \begin{subfigure}[t]{0.24\linewidth}
        \includegraphics[width=\linewidth]{\rootnaturev2 Restormer/Restormer_1-2_71.pdf}
        \caption{\small Restormer (PSNR: 23.50)}
    \end{subfigure} \hspace{-5pt}
    \begin{subfigure}[t]{0.24\linewidth}
        \includegraphics[width=\linewidth]{\rootnaturev2 RSIRR/RSIRR_1-2_71.pdf}
        \caption{\small RSIRR (PSNR: 22.69)}
    \end{subfigure} \hspace{-5pt}
    \begin{subfigure}[t]{0.24\linewidth}
        \includegraphics[width=\linewidth]{\rootnaturev2 PromptRR/PromptRR_1-2_71.pdf}
        \caption{\small Ours (PSNR: 23.57)}
    \end{subfigure} \hspace{-5pt}
    \begin{subfigure}[t]{0.24\linewidth}
        \includegraphics[width=\linewidth]{\rootnaturev2 GT/GT_1-2_71.pdf}
        \caption{\small GT}
    \end{subfigure}
      \end{minipage}
}
 \vspace{8pt}
    % \caption{Visual comparison on \textit{Nature}~\cite{Li_2020_CVPR} dataset.\textbf{Zoom-in for better details}.}
    \caption{Visual comparison on \textit{Nature}~\cite{Li_2020_CVPR} dataset. Compared with other methods, our PromptRR effectively removes reflections while preserving the fine details in the restored images. \textbf{Zoom-in for better details}.}
    \label{fig:results_nature}
\end{figure*}

\section{Experiments}\label{sec:experiments}

\begin{figure*}[t]
 \begin{center}
   \begin{overpic}[width=\textwidth]{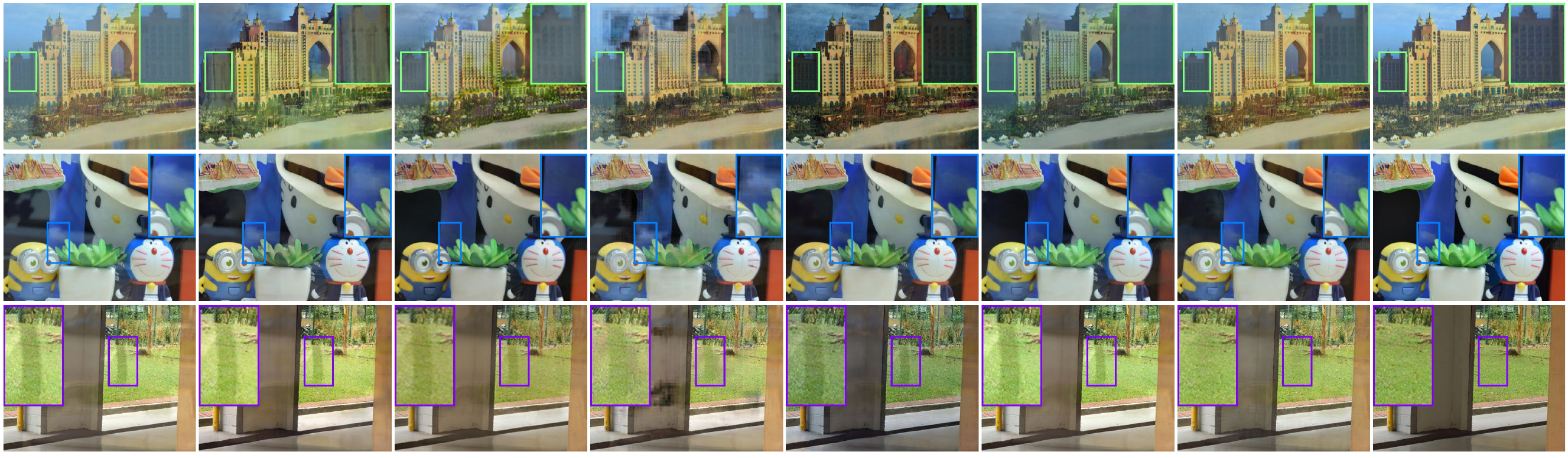}
     \put(5,-1.5){Input}
   \put(16.5,-1.5){CL21}
   \put(28.5,-1.5){HG21}
   \put(40,-1.5){Uformer}
   \put(52.5,-1.5){Restormer}
   \put(66,-1.5){RSIRR}
   \put(77,-1.5){PromptRR}
   \put(91.5,-1.5){GT}
  \end{overpic}
   \vspace{6pt}
  \caption{Visual comparison on \textit{$SIR^{2}$} dataset. The images are from three subsets: \textit{Postcard} (1st row), \textit{SolidObject} (2nd row), and \textit{WildScene} (3rd row). }
    % \caption{Visual comparison on \textit{$SIR^{2}$} dataset. The images are from three subsets: \textit{Postcard} (1st row), \textit{SolidObject} (2nd row), and \textit{WildScene} (3rd row). In contrast, our PromptRR removes more undesired reflections and produces clearer images (see bounding boxes).}
 	\label{fig:results_SIR2}
 	\end{center}
 % \vspace{-2mm}
 \end{figure*}

\def \rootrealv2 {images/datasets-v2/real/origin_img_with_box/}

\begin{figure*}[ht]
    \centering
        \resizebox{1\textwidth}{!}{  % Scale the figure, adjust as needed
    \begin{minipage}{\textwidth}
        \centering
    \begin{subfigure}{0.24\linewidth}
        \includegraphics[width=\linewidth]{\rootrealv2 input/input_89.pdf}
        \caption{Input (PSNR: 18.06)}
    \end{subfigure} \hspace{-5pt}
    \begin{subfigure}{0.24\linewidth}
        \includegraphics[width=\linewidth]{\rootrealv2 WY19/WY19_89.pdf}
        \caption{WY19 (PSNR: 19.61)}
    \end{subfigure} \hspace{-5pt}
    \begin{subfigure}{0.24\linewidth}
        \includegraphics[width=\linewidth]{\rootrealv2 WT19/WT19_89.pdf}
        \caption{WT19 (PSNR: 7.60)}
    \end{subfigure} \hspace{-5pt}
    \begin{subfigure}{0.24\linewidth}
        \includegraphics[width=\linewidth]{\rootrealv2 LY20/LY20_89.pdf}
        \caption{LY20 (PSNR: 20.36)}
    \end{subfigure}
    \\  % New row
    \begin{subfigure}{0.24\linewidth}
        \includegraphics[width=\linewidth]{\rootrealv2 ZS21/ZS21_89.pdf}
        \caption{ZS21 (PSNR: 17.72)}
    \end{subfigure} \hspace{-5pt}
    \begin{subfigure}{0.24\linewidth}
        \includegraphics[width=\linewidth]{\rootrealv2 CL21/CL21_89.pdf}
        \caption{CL21 (PSNR: 20.87)}
    \end{subfigure} \hspace{-5pt}
    \begin{subfigure}{0.24\linewidth}
        \includegraphics[width=\linewidth]{\rootrealv2 HG21/HG21_89.pdf}
        \caption{HG21 (PSNR: 22.45)}
    \end{subfigure} \hspace{-5pt}
    \begin{subfigure}{0.24\linewidth}
        \includegraphics[width=\linewidth]{\rootrealv2 Uformer/Uformer_89.pdf}
        \caption{Uformer (PSNR: 18.37)}
    \end{subfigure}
    \\  % New row
    \begin{subfigure}{0.24\linewidth}
        \includegraphics[width=\linewidth]{\rootrealv2 Restormer/Restormer_89.pdf}
        \caption{Restormer (PSNR: 18.67)}
    \end{subfigure} \hspace{-5pt}
    \begin{subfigure}{0.24\linewidth}
        \includegraphics[width=\linewidth]{\rootrealv2 RSIRR/RSIRR_89.pdf}
        \caption{RSIRR (PSNR: 24.77)}
    \end{subfigure} \hspace{-5pt}
    \begin{subfigure}{0.24\linewidth}
        \includegraphics[width=\linewidth]{\rootrealv2 PromptRR/PromptRR_89.pdf}
        \caption{Ours (PSNR: 26.15)}
    \end{subfigure} \hspace{-5pt}
    \begin{subfigure}{0.24\linewidth}
        \includegraphics[width=\linewidth]{\rootrealv2 GT/GT_89.pdf}
        \caption{GT}
    \end{subfigure} 
    \end{minipage}
}
    \caption{Visual comparison on \textit{Real}~\cite{zhang2018single} dataset. \textbf{Zoom-in for better details}.}
    \label{fig:results_real}
\end{figure*}
\subsection{Experimental settings}
%%% Tao Wang revised 
\noindent \textbf{Datasets and Metrics}. We conduct experiments on popular benchmarks. Following previous works~\cite{hu2021trash,song2023robust}, we train our networks on synthetic and real-world datasets. For the synthetic training data, we adopt the PASCAL VOC dataset~\cite{Everingham2010} to synthesize $7,643$ images using the data-generation method described in \cite{fan2017generic}. For the real-world data, we adopt $90$ real-world training images from~\cite{zhang2018single}. In the testing phase, we evaluate models on three commonly used real-world datasets, Nature~\cite{Li_2020_CVPR}, $SIR^{2}$~\cite{Wan_2017_iccv}, and Real dataset~\cite{zhang2018single}. The $SIR^{2}$ dataset comprises three subsets, namely \textit{PostCard}, \textit{SolidObject}, and \textit{WildScene}. In addition, PSNR~\cite{huynh2008scope} and SSIM~\cite{wang2004image} are used as the metrics for performance evaluation.

\vspace{1mm}
\noindent \textbf{Comparison SIRR Methods}. We compare the proposed PromptRR with six CNN-based SIRR methods (WY19~\cite{Wei_2019_CVPR}, WT19~\cite{Wen_2019_CVPR}, LY20~\cite{Li_2020_CVPR},  ZS21~\cite{Zheng_2021_CVPR}, CL21~\cite{Chang2021_wacv}, HG21~\cite{hu2021trash} and  
one recent Transformer-based SIRR method (RSIRR~\cite{song2023robust}), and two general Transformer-based image restoration methods (Uformer~\cite{wang2022uformer} and Restormer~\cite{zamir2022restormer}).

\vspace{1mm}
\noindent \textbf{Training Details}. Our PromptRR framework contains two stages: prompt pre-training and prompt generation and restoration. PromptFormer follows a hierarchical transformer-based architecture with transformer block configurations set to ${N_{0}, N_{1}, N_{2}, N_{3}}={4,6,6,8}$, where each stage processes features at different resolutions. The number of attention heads is configured as ${1, 2, 4, 8}$ to facilitate effective multi-scale feature learning, and the initial dimension of the channel C is set to $48$. Specifically, in the prompt pre-training stage, we employ the AdamW optimizer to train our FPE and PromptFormer models. The training is performed with a batch size of $8$ and a patch size of 128 for $200,000$ iterations. The learning rate is set to $1\times10^{-4}$. For the prompt generation and restoration stage, we utilize the Adam optimizer with a batch size of 8 and a patch size of 128. In the initial phase, the dual-diffusion model is trained for $20,000$ iterations. Subsequently, we jointly train the dual-diffusion model and PromptFormer for $280,000$ iterations. The learning rate is set to $1\times10^{-4}$. For the diffusion modes, the total timestep $T$ is set to 4 to ensure an effective denoising process.

\subsection{Comparisons Results}
%%%%% write by tao wang_over
\noindent \textbf{Quantitative Results}. Table~\ref{tab:quanti_res} presents the quantitative comparisons of our method with the SOTA SIRR methods on commonly used real-world dataset datasets. Recent image restoration methods Restormer~\cite{zamir2022restormer} and Uformer~\cite{wang2022uformer} adopt an end-to-end network to achieve image restoration. 
However, they do not consider explicitly reflection-conditional guidance to help the restoration process. Thus, these methods do not effectively remove reflection. Although RSIRR~\cite{song2023robust} utilizes 
cross-scale attention, multi-scale, and adversarial mechanisms to address SIRR, its performance is still unsatisfactory. In contrast, our PromptRR can explore frequency prompts to guide the process of reflection removal. As shown in Table~\ref{tab:quanti_res}, our PromptRR achieves the best performance in terms of PSNR and SSIM on all real-world datasets, where the average PSNR value is $1.4$ dB higher than the second-place method RSIRR~\cite{song2023robust} on \textit{$SIR^{2}$} dataset.

%%%%% written by tao wang_over 
\vspace{1mm}
\noindent \textbf{Qualitative Results}. We also visually compared the reflection removal results of five SOTA models and our PromptRR. Specifically, as demonstrated in Fig.~\ref{fig:results_nature}, Fig.~\ref{fig:results_SIR2}, and Fig.~\ref{fig:results_real}, we present visual comparisons on the employed datasets. Our method effectively removes reflections, yielding images resembling the ground truth. In Fig.~\ref{fig:results_nature}, while the compared methods demonstrate effective restoration, noticeable striped shadow artifacts remain in the region highlighted by the blue boxes, particularly in the second-best method, RSIRR~\cite{song2023robust}. In contrast, our approach eliminates these residual reflections and preserves fine details, resulting in more visually faithful and high-quality restored images.
Fig.~\ref{fig:results_SIR2} (the third-row purple boxes) illustrates the failure of compared methods to eliminate shadow effects, in contrast to our approach, which can remove reflections to produce more pleasing glasses. In Fig.~\ref{fig:results_real}, most methods struggle to clarify toy structures and textures under severe degradation, while PrompRR produces the restored images with better structural fidelity. In addition, we present the PSNR results, and our method obtains the highest values in PSNR. As a result, our method shows its superiority compared to other existing methods. More qualitative
results can be found in our supplemental material.

\subsection{Ablation Study}
%%%%% write by tao wang_over 
We conduct a comprehensive analysis of the proposed method to understand how it tackles the issue of SIRR and to show the impact of its primary components. 
To ensure fair comparisons in the ablation studies, we only train all model variants on the training set for $100,000$ iterations. We evaluate the models on \textit{Real} testing set in terms of PSNR and SSIM.

\vspace{1mm}
\noindent \textbf{Effectiveness of PIIM}. 
%%%%% write by tao wang_over
To verify whether our PIIM facilitates reflection removal, we conduct ablation studies by setting two other variants: (1) removing the prompt interaction stage in PIIM from the model and (2) removing all PIIM from the model, respectively. 
% We train these two variant models using the same settings as our method on the training set. 
Table~\ref{table_ablation_PIIM} presents that our full model with the proposed PIIM exhibits superior performance compared to other models, where the PSNR and SSIM values surpass that of the model without PIIM by $3.79$ dB and $0.0856$. Moreover, the model incorporating the prompt interaction stage in PIIM achieves higher PSNR and SSIM values. Specifically, the PSNR value is $0.53$ dB higher than the model without the prompt interaction stage in PIIM. This suggests that using the prompt interaction in PIIM aids in refining and reinforcing prompts for better reflection removal.
% \begin{table}[t] 
% 	\centering
%     \caption{Effectiveness of the proposed PIIM.}
% 	% \vspace{-3mm}
% 	\resizebox{0.9\columnwidth}{!}{
% 	\begin{tabular}{cccc}
% 	\hline 
% 	Models       &  w/o Interaction    & w/o PIIM& w/ PIIM (Ours)                     \\ \hline
% 	PSNR/SSIM  &23.51/0.8011 &20.25/0.7247  & \textbf{24.04}/\textbf{0.8103}  \\ \hline 
% \end{tabular}
% }	
 
%  \label{table_ablation_PIIM}
% 	% \vspace{-2mm}
% \end{table}

\begin{table}[t]\small
	\centering
	% \vspace{-3mm}
    \caption{Effectiveness of the proposed PIIM.}
	% \resizebox{0.9\columnwidth}{!}{
	\begin{tabular}{cccc}
	\hline 
	Models & Interaction & PIIM  & PSNR/SSIM   \\ \hline
	(1)    &        & $\checkmark$  & 23.51/0.8011 \\
	(2)    &        &      & 20.25/0.7247 \\
	Ours   &   $\checkmark$    & $\checkmark$  & 24.04/0.8103 \\ \hline 
    \end{tabular}
	% }	
 
 \label{table_ablation_PIIM}
	% \vspace{-4mm}
\end{table}

% \begin{wraptable}{r}{0.5\textwidth}
% % \begin{table}[t]\small
%         \tabcolsep=1pt % 调整列间距
%   \renewcommand\arraystretch{1.1}
% 	\centering
% 	\caption{Effectiveness of the proposed PIIM.}
% 	% \vspace{-3mm}
% 	\resizebox{0.5\columnwidth}{!}{
% 	\begin{tabular}{cccc}
% 	\hline 
% 	Models       &  w/o Interaction    & w/o PIIM& w/ PIIM (Ours)                     \\ \hline
% 	PSNR/SSIM  &23.51/0.8011 &20.25/0.7247  & \textbf{24.04}/\textbf{0.8103}  \\ \hline 
% \end{tabular}
% }	
% 	\label{table_ablation_PIIM}
% 	% \vspace{-8mm}
% % \end{table}

% \end{wraptable}

\begin{table}[t]\small
	\centering
	% \vspace{-3mm}
    \caption{Effectiveness of the prompt for reflection removal.}
	% \resizebox{0.9\columnwidth}{!}{
	\begin{tabular}{cccccc}
	\hline 
	Models & MSA & FFN & PMSA & PFFN & PSNR/SSIM   \\ \hline
	(a)    &   $\checkmark$     & $\checkmark$    &        &     & 20.25/0.7247 \\
	(b)    & $\checkmark$       &    &       & $\checkmark$    & 23.29/0.8010 \\
	(c)    &       & $\checkmark$    & $\checkmark$       &        & 22.48/0.7973 \\
	Ours   &     &  & $\checkmark$      &$\checkmark$    & \textbf{24.04}/\textbf{0.8103} \\ \hline 
    \end{tabular}
	% }	
 
 \label{table_ablation_prompt}
	% \vspace{-4mm}
\end{table}
% \begin{table}[t] \small
% 	\centering
% 	\caption{Ablation study on prompt generation using DMs.}
% 	% \vspace{-3mm}
% 	\resizebox{0.8\columnwidth}{!}{
% 	\begin{tabular}{ccc}
% 	\hline 
% 	PromptRR      & w/o DMs   & w/ DMs                     \\ \hline
% 	PSNR/SSIM   & 19.95/0.7328 &\textbf{24.04}/\textbf{0.8103}    \\ \hline 
% \end{tabular}
% 	}	
% 	\label{table_ablation_DMs}
% 	% \vspace{-6mm}
% \end{table}

\begin{table}[t] \small
	\centering
	% \vspace{-3mm}
     \caption{Ablation study on prompt generation using DMs.}
	% \resizebox{0.7\columnwidth}{!}{
	\begin{tabular}{ccc}
	\hline 
	PromptRR      & w/o DMs   & w/ DMs                     \\ \hline
	PSNR/SSIM   & 19.95/0.7328 &\textbf{24.04}/\textbf{0.8103}    \\ \hline 
\end{tabular}
	% }	

 \label{table_ablation_DMs}
	% \vspace{-4mm}
\end{table}

\begin{figure}[!t]
	\centering 	
	\begin{subfigure}[t]{0.32\columnwidth}
		\centering
		\includegraphics[width=0.96\columnwidth]{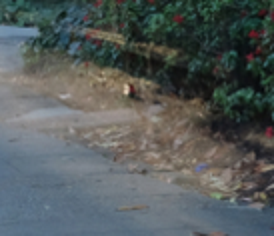}
		\caption{Input}
	\end{subfigure}
	\begin{subfigure}[t]{0.32\columnwidth}
		\centering
		\includegraphics[width=0.96\columnwidth]{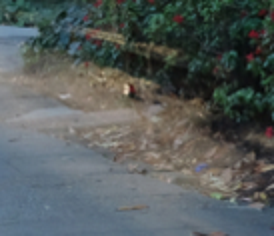}
		\caption{Model (a)}
	\end{subfigure}
	\begin{subfigure}[t]{0.32\columnwidth} 
	\centering 
	\includegraphics[width=0.96\columnwidth]{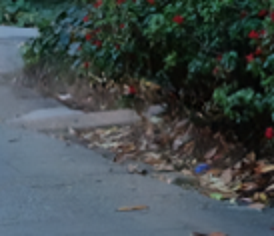}
	\caption{Model (b)}
    \end{subfigure} \\ 	
    \begin{subfigure}[t]{0.32\columnwidth}
    \centering
    \includegraphics[width=0.96\columnwidth]{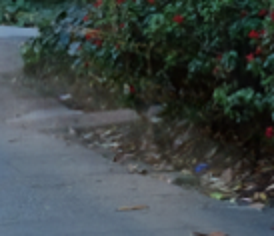}
    \caption{Model (c)}
    \end{subfigure}
        \begin{subfigure}[t]{0.32\columnwidth}
    \centering
    \includegraphics[width=0.96\columnwidth]{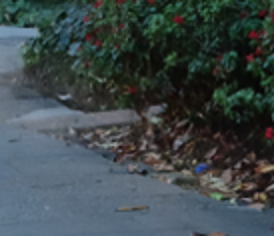}
    \caption{Ours}
    \end{subfigure}
        \begin{subfigure}[t]{0.32\columnwidth}
    \centering
    \includegraphics[width=0.96\columnwidth]{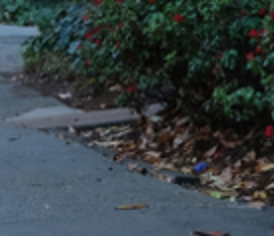}
    \caption{GT}
    \end{subfigure}
	% \vspace{-3mm}
	\caption{Visual comparison between our method and variants.}
	\label{fig_ablation_prompt}
	% \vspace{-2mm}
\end{figure}

%%%%% write by tao wang_over 
\vspace{1mm}
\noindent \textbf{Effectiveness of Prompt}.Our method employs our transformer-based prompt block as the basic unit for constructing our PromptFormer. Within the prompt block, we introduce the frequency prompt into the plain multi-head self-attention (MAS) and plain feed-forward network (FFN)~\cite{zamir2022restormer} to construct the core modules PMSA and PFFN. To better demonstrate the effectiveness of the prompt used in PMSA and PFFN, we conduct an ablation study that considers the position of the prompt used in the prompt block. As shown in Table~\ref{table_ablation_prompt}, we build three baselines with different positions of the prompt used within the prompt block: (a) without using any frequency prompt in both MSA and FFN, (b) using the frequency prompt in FFN, and (c) using the frequency prompt in MSA. Table~\ref{table_ablation_prompt} and Fig.~\ref{fig_ablation_prompt} show the quantitative and qualitative results.
The models ' performance is improved when using the prompt in MSA (c) or FFN (b). Compared to other baselines, our model yields the best performance after introducing the prompt into MSA and FFN, which validates the effectiveness of the prompt. Fig.~\ref{fig_ablation_prompt} further shows that using the prompts to guide the model can facilitate reflection removal.

%
% \begin{figure}[!t]
% 	\centering
% 	\setlength{\tabcolsep}{0pt} % Reduce spacing between subfigures
% 	\renewcommand{\arraystretch}{0} % Remove extra spacing around captions
% 	\begin{tabular}{cccccc} % Create a grid of subfigures
% 		\begin{subfig}[t]{0.16\linewidth}
% 			\centering
% 			\includegraphics[width=\textwidth]{images/table3_imgs/input_25_magnifier_0.png}
% 		\end{subfig} &
% 		\begin{subfig}[t]{0.16\linewidth}
% 			\centering
% 			\includegraphics[width=\textwidth]{images/table3_imgs/ValSet_real20_a_25_magnifier_0.png}
% 			\caption{Model (a)}
% 		\end{subfig} &
% 		\begin{subfig}[t]{0.16\linewidth}
% 			\centering
% 			\includegraphics[width=\textwidth]{images/table3_imgs/ValSet_real20_b_25_magnifier_0.png}
% 			\caption{Model (b)}
% 		\end{subfig} &
% 		\begin{subfig}[t]{0.16\linewidth}
% 			\centering
% 			\includegraphics[width=\textwidth]{images/table3_imgs/ValSet_real20_c_25_magnifier_0.png}
% 			\caption{Model (c)}
% 		\end{subfig} &
% 		\begin{subfig}[t]{0.16\linewidth}
% 			\centering
% 			\includegraphics[width=\textwidth]{images/table3_imgs/ValSet_real20_d_25_magnifier_0.png}
% 			\caption{Ours}
% 		\end{subfig} &
% 		\begin{subfig}[t]{0.16\linewidth}
% 			\centering
% 			\includegraphics[width=\textwidth]{images/table3_imgs/gt_25_magnifier_0.png}
% 			\caption{GT}
% 		\end{subfig}
% 	\end{tabular}
% 	\caption{Visual comparison between our method and variants.}
% 	\label{fig:ablation_prompt}
% \end{figure}

\begin{figure}[!t]
	\centering 	
	\begin{subfigure}[t]{0.24\columnwidth}
		\centering
		\includegraphics[width=\columnwidth]{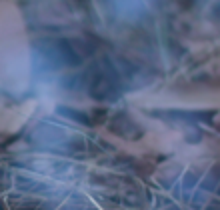}
		\caption{Input}
	\end{subfigure}
	\begin{subfigure}[t]{0.24\columnwidth}
		\centering
		\includegraphics[width=\columnwidth]{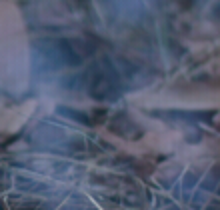}
		\caption{w/o DMs}
	\end{subfigure}
	\begin{subfigure}[t]{0.24\columnwidth}
	\centering
	\includegraphics[width=\columnwidth]{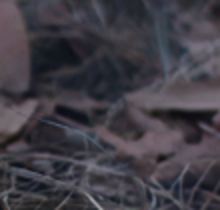}
	\caption{w/ DMs}
    \end{subfigure}	
    \begin{subfigure}[t]{0.24\columnwidth}
    \centering
    \includegraphics[width=\columnwidth]{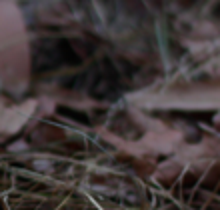}
    \caption{GT}
    \end{subfigure}
	% \vspace{-3mm}
	\caption{Ablation qualitative comparison for DMs for prompt generation.}
	\label{fig_ablation_DMs}
	\vspace{-3mm}
\end{figure}

% \begin{figure}[!t]
% 	\centering 	
% 	\parbox[t]{0.10\textwidth}{%
% 		\centering
% 		\includegraphics[width=0.10\columnwidth]{images/table4_imgs/input_107_magnifier_0.png}
% 		\caption{Input}
% 	}
% 	\parbox[t]{0.10\textwidth}{%
% 		\centering
% 		\includegraphics[width=0.10\columnwidth]{images/table4_imgs/ValSet_real20_ab1_107_magnifier_0.png}
% 		\caption{w/o DMs}
% 	}
%         \parbox[t]{0.10\textwidth}{%
%         \centering
%         \includegraphics[width=0.10\columnwidth]{images/table4_imgs/ValSet_real20_ours_ab_107_magnifier_0.png}
%         \caption{w/ DMs}
%         }%
%         \parbox[t]{0.10\textwidth}{%
%         \centering
%         \includegraphics[width=0.10\columnwidth]{images/table4_imgs/gt_107_magnifier_0.png}
%         \caption{GT}
%         }
% 	% \vspace{-3mm}
% 	\caption{Ablation qualitative comparison for DMs for prompt generation.}
% 	\label{fig_ablation_DMs}
% 	\vspace{-3mm}
% \end{figure}
% ablation study for models with or without frequency prompts could applied here.  
\vspace{1mm}
\noindent \textbf{Effectiveness of DMs for Prompt Generation}. 
We further demonstrate the effectiveness of DMs for prompt generation. In Table~\ref{table_ablation_DMs}, 
``w/o DMs'' means using a CNN network to replace the DMs for prompt generation in our PromptPR. ``DMs'' refers to our PromptRR using diffusion models to generate prompts for guiding reflection removal. The results are listed in Table.~\ref{table_ablation_DMs} and Fig.~\ref{fig_ablation_DMs}. It shows that using DMs for prompt generation achieves the best performance of reflection removal in PSNR and SSIM and the best visual effects. Compared with the model without DMs as prompt generators, the improvement gains in PSNR and SSIM are  $4.09$ dB and $0.0775$, respectively. From Fig.~\ref{fig_ablation_DMs}, we observe that using DMs as the prompt generator to generate prompts, our PromptRR achieves the closest visual results to the ground truth. 
In contrast, the model without DMs fails to remove reflections from the image. Thus, the results show that DMs are powerful tools for high-quality prompt generation.

\section{Conclusion}
\label{sec:conclusion}

In this paper, we propose a novel Prompt-guided Reflection Removal Framework (PromptRR), which utilizes diffusion models (DMs) as frequency prompt generators to facilitate effective reflection removal. We decompose the reflection removal process into two stages: prompt generation and prompt-guided restoration. To generate high-quality frequency prompts, we introduce a prompt pre-training strategy, wherein DMs are trained to generate these prompts. Once trained, the DMs serve as frequency prompt generators to first produce the prompts, which are then embedded into the PromptFormer network for enhanced reflection removal. Experimental results demonstrate that PromptRR outperforms state-of-the-art methods.

\bibliographystyle{IEEEtran}

\bibliography{main}

\end{document}